\documentclass{article}

\usepackage{PRIMEarxiv}

\usepackage[utf8]{inputenc} 
\usepackage[T1]{fontenc}    
\usepackage{hyperref}       
\usepackage{url}            
\usepackage{booktabs}       
\usepackage{amsfonts}       
\usepackage{nicefrac}       
\usepackage{microtype}      
\usepackage{lipsum}
\usepackage{fancyhdr}       
\usepackage{graphicx}       
\graphicspath{{media/}}     

\usepackage{wrapfig}
\usepackage{multirow}
\usepackage{subfigure}
\usepackage{booktabs}
\usepackage{amsmath}
\usepackage{amsfonts}
\usepackage{color, colortbl}
\usepackage{xcolor}
\usepackage{pgfplots}
\usepackage{bm}

\newcommand{\myparagraph}[1]{\smallskip\noindent\textbf{#1}\space}

\definecolor{red}{HTML}{E41A1C}
\definecolor{green}{HTML}{4DAF4A}
\definecolor{blue}{HTML}{377EB8}

\newcommand{\squishlist}{
	\begin{list}{$\bullet$}
		{ \setlength{\itemsep}{0pt}      \setlength{\parsep}{3pt}
			\setlength{\topsep}{3pt}       \setlength{\partopsep}{0pt}
			\setlength{\leftmargin}{1.5em} \setlength{\labelwidth}{1em}
			\setlength{\labelsep}{0.5em} } }
	
	\newcommand{\squishlisttwo}{
		\begin{list}{$\bullet$}
			{ \setlength{\itemsep}{0pt}    \setlength{\parsep}{0pt}
				\setlength{\topsep}{0pt}     \setlength{\partopsep}{0pt}
				\setlength{\leftmargin}{2em} \setlength{\labelwidth}{1.5em}
				\setlength{\labelsep}{0.5em} } }
		
		\newcommand{\squishend}{	\end{list}  }

\definecolor{PastelRed}{rgb}{0.98, 0.502, 0.447}
\definecolor{PastelBlue}{rgb}{0.502, 0.694, 0.827}

\def\R{\mathbb{R}}
\def\N{\mathbb{N}}
\newcommand{\T}[1]{\bm{\mathcal{#1}}}
\newcommand{\V}[1]{\mathbf{#1}}
\newcommand{\aspas}[1]{``{#1}''}

\newcommand\crule[3][black]{\textcolor{#1}{\rule{#2}{#3}}}

\title{TensorAnalyzer: Identification of Urban Patterns in Big Cities using Non-Negative Tensor Factorization
}

\author{
  Jaqueline Silveira, Germain Garc\'ia, Afonso Paiva, Luis Gustavo Nonato \\
  ICMC-USP \\
  São Carlos\\
  \texttt{\{alva.jaque, germaingarcia,apneto\}@usp.br} \\
  \texttt{gnonato@icmc.usp.br} \\
   \And
  Marcelo Nery, Sergio Adorno \\
  NEV-USP \\
  São Paulo\\
  \texttt{mbnery@gmail.br} \\
  \texttt{marsadorno@usp.br} \\
}

\begin{document}
\maketitle

\begin{abstract}
Extracting relevant urban patterns from multiple data sources can be difficult using classical clustering algorithms since we have to make a suitable setup of the hyperparameters of the algorithms and deal with outliers. It should be addressed correctly to help urban planners in the decision-making process for the further development of a big city. For instance, experts' main interest in criminology is comprehending the relationship between crimes and the socio-economics characteristics at specific georeferenced locations. In addiction, the classical clustering algorithms take little notice of the intricate spatial correlations in georeferenced data sources.
This paper presents a new approach to detecting the most relevant urban patterns from multiple data sources based on tensor decomposition. Compared to classical methods, the proposed approach's performance is attested to validate the identified patterns' quality. The result indicates that the approach can effectively identify functional patterns to characterize the data set for further analysis in achieving good clustering quality. Furthermore, we developed a generic framework named TensorAnalyzer, where the effectiveness and usefulness of the proposed methodology are tested by a set of experiments and a real-world case study showing the relationship between the crime events around schools and students performance and other variables involved in the analysis.
\end{abstract}

\keywords{crime analysis \and school data \and non-negative tensor decomposition \and information visualization}

\section{Introduction}
\begin{figure*}[th]
   \centering
	\includegraphics[width=\linewidth]{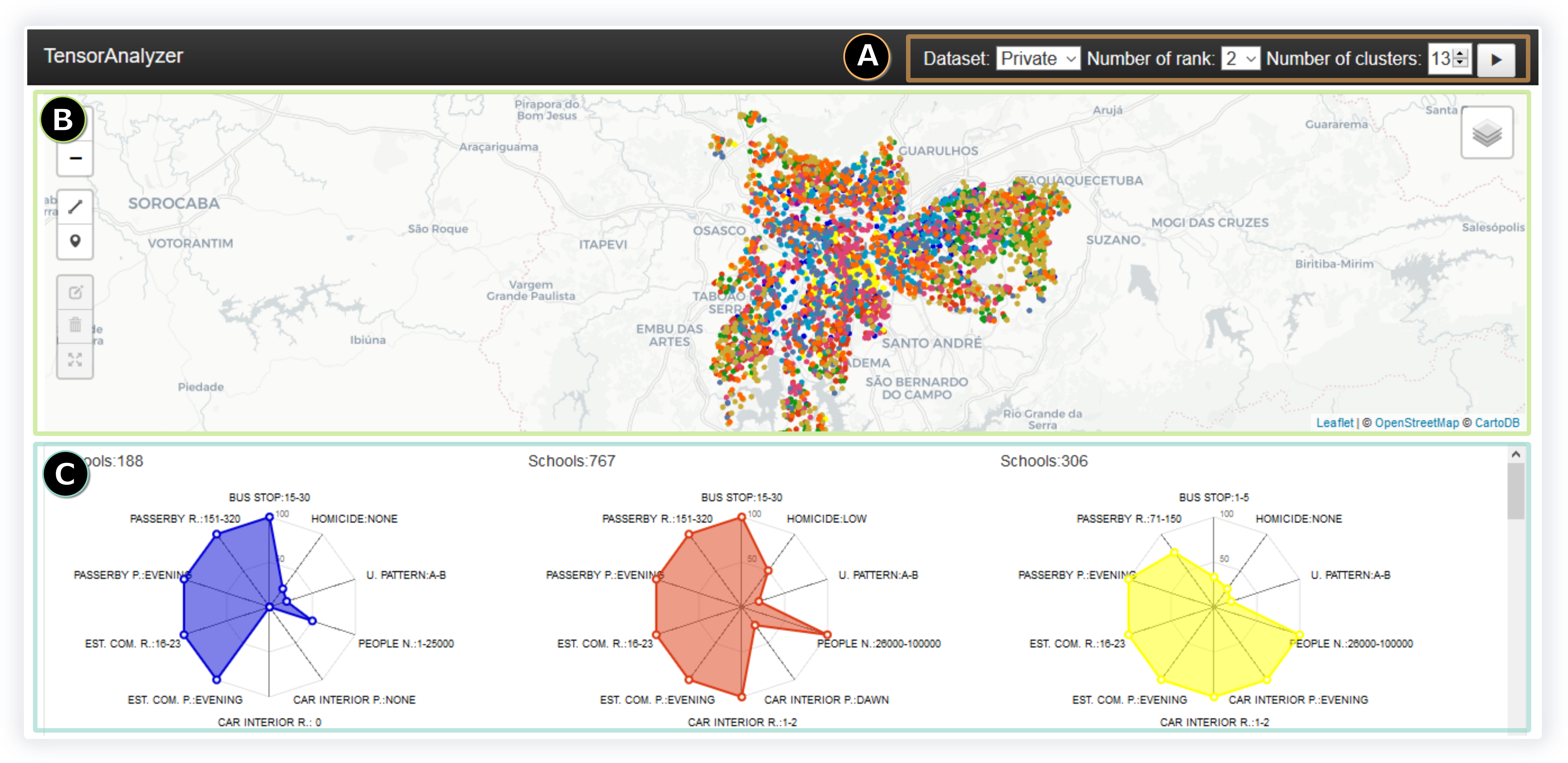}
	\caption{TensorAnalyzer System: the visualization of patterns enables the understanding of the relationship between crimes and other variables involved in the analysis. Our visual tool comprises a \emph{Control Menu}~(A), \emph{Map View}~(B), and \emph{Patterns View}~(C). }
	\label{fig:teaser}
\end{figure*}

In Spatial Data Analysis, one of the main concepts is Point Pattern Analysis, which seeks to understand the relationship between the location of events and the response to the distribution of these locations. More specifically, the interest is to verify if a point pattern follows some systematic clustering process or if it follows a random process \cite{Yamada2010}. The information obtained from these processes may allow us to acquire some initial insights into a given phenomenon. For instance, extraction of urban patterns in big cities is an emerging application that has helped public policymakers understand the relationship between crime variables, urban infrastructure, and socio-economic indicators and how it affects the population's daily lives in specific city regions.

Identifying urban patterns in multiple data sources can be not straightforward \cite{Picornell2019,zhao2015explaining,chen2003,DeBruin2006}. Because it involves manipulating and combining different data sources to group places or regions that share similar characteristics, even in the presence of a hidden correlation between these variables. A possible solution for this issue would be clustering algorithms~\cite{bousema2010,cook2011serological,huang2006,campelo2016,Malathi2011,Joshi2017}. However, the wrong setting of the hyperparameters of the classical clustering algorithms can lead to an inaccurate extraction of the patterns due to a noise behavior in real databases~\cite{gupta2019dealing}.In addition, very little attention has been paid to the dependency between different data sources georeferenced during the clustering process. We argue that capturing the correlations between different data sources georeferenced is of great help to improve the final results. For instance,  manytheft crimes are drug-related~\cite{Fnu2018} because the persons who have become addicted to street drugs generally commit theft to fund and support their drug habit. Therefore, developing a sophisticated and robust data science tool, which has not too hyper-parameters to set, capable of extracting relevant urban patterns, taking into account the intricate correlations between the georeferenced data sources, is crucial in this work.

This paper introduces a novel methodology based on data tensor modeling and tensor factorization to find the most relevant urban patterns around target locations in a big city. This strategy allows combining different georeferenced data sources into a tensor, while the tensor's pattern extraction relies on \emph{Non-Negative Tucker Decomposition}~\cite{cichocki2009nonnegative}.
Our approach is robust against noise and capable of highlighting the correlation between the urban characteristics.
The proposed methodology's effectiveness and usefulness are tested by a set of experiments and a real-world case study showing the relationship between the urban patterns provided by our approach and the students' performance in São Paulo city, the largest city in Latin America. Besides, we developed a visualization-assisted tool called by TensorAnalyzer, to allow visual exploration and analysis of georeferenced data regarding urban patterns.

In summary, the main contributions of this paper are: 

\begin{itemize}
	\item A new approach to identify the most relevant urban patterns based on tensor modeling and decomposition techniques;
	\item TensorAnalyzer, a visualization-assisted tool that operates on target locations handling multiple data sources to uncover relevant information about the regions as a whole;
	\item A comprehensive validation study through qualitative and quantitative experiments in synthetic and real data sets. These results show the proposed approach's capability and potential, revealing interesting phenomena and events from massive data.
\end{itemize}

\section{Related Work}

Since our approach extracts urban patterns from georeferenced data using tensor decomposition, to better contextualize our approach, we organize the existing methods for pattern extraction from georeferenced data and applications of tensor decomposition for data processing.

\subsection*{Pattern extraction from georeferenced data}
Georeferenced data, also known as spatial, geographic, or geospatial data, are the essential pieces of information necessary to identify the geographic location of phenomena on the surface of the Earth ~\cite{Anthony1996}. In general, georeferenced data is indexed by geographic coordinates (points referenced by latitude and longitude) that allow the application of spatial pattern extraction techniques to extract valuable information from the data. The following is an overview of the main fields investigated in the literature that seek to find patterns considering several georeferenced data sources.

Point pattern analysis has been a growing interest in identifying and analyzing forest fire sites. For example, \cite{podur2003spatial} studied lightning patterns that caused fires in Ontario from 1976 to 1998. The authors used K-function to evaluate the clustering and the smoothing of the kernel density in order to provide graphical representations of the cluster. \cite{wotton2005lightning} developed a logistic model to estimate the probability of occurrences of fires caused by lightning, considering meteorological variables and indices related to fire risk. Point pattern analysis has also been widely applied in the field of epidemiology in order to understand the relationship between health events and aspects related to individual characteristics (e.g., genetics, behavior, and demographic) and contextual factors (e.g., the socio-economic conditions of the surroundings and the physical environment), in which clustering algorithms are applied to detect spatial groups of diseases ~\cite{Werneck2008,pullan2012spatial}. In this context, ~\cite{bousema2010} used cluster statistics to show that clusters of HIV-positive individuals can be used to identify hotspots with a high incidence of malaria in Tanzania. ~\cite{cook2011serological} used spatial scanning to explore the clustering of malaria infection in different age groups on Bioko Island in Equatorial Guinea.

Recently, we have been witnessing an emerging class of communication platforms, such as Facebook, Twitter, FourSquare, and Flickr, that allow users to publish georeferenced content ranging from short status messages to videos. Some studies have been developed using this content information to extract patterns from the data. For example, \cite{huang2006} developed a tool that uses spatial clustering techniques to extract data patterns from georeferenced mobile devices. \cite{Yuan2011} focused on extracting activities and human mobility based on data from mobile devices. \cite{zhao2015explaining} conducted a statistical analysis that aims to identify general spatial patterns in the occurrence of flood-related tweets that may be associated with the proximity and severity of flood events. \cite{campelo2016} presented an approach based on the clustering algorithm DBSCAN~\cite{Ester1996} to extract mobility patterns from Twitter messages and then analyze their correlation with demographic, economic, and social data. \cite{Picornell2019} presented a new methodology based on clustering algorithms to estimate population dynamics from cell phone data.

In the context of crimes, data are becoming increasingly available to the population, which has called researchers’ attention to try to understand the criminal patterns of a given location. Crime records are generally georeferenced data; therefore, several approaches have been proposed to extract spatial patterns from these data. For instance, the COPLINK~\cite{chen2003} system is recognized as one of the successful implementations of the clustering technique for criminal data mining, in which the concept of space is applied to identify relationships between suspects and victims~\cite{Hou1994}. \cite{Brown1998} developed a framework named ReCAP (Regional Crime Analysis Program) that combines multiple data sources to extract patterns. \cite{chen2004} developed a general tool that allows discovering associations, identifying patterns, and making predictions. ~\cite{DeBruin2006} described a tool capable of extracting important features from the georeferenced database and creating digital profiles for all aggressors. Their method compares all individuals in these profiles using a new distance measure and groups them together. \cite{Malathi2011} developed a tool to detect crime patterns based on data mining techniques such as clustering and classification. \cite{Aljanabi2011} proposed a tool for analyzing criminal data that focuses on identifying patterns and trends that uses decision trees to classify the data together with clustering algorithms. \cite{Joshi2017} employed the \textit{k-means}~\cite{Jin2010} algorithm to extract criminal patterns from large georeferenced data sets. 


The works described in this section aimed at identifying spatial patterns in georeferenced data and most of them use standard clustering algorithms for this purpose. However, these algorithms require user tuning parameters, a complicated task when the user does not have prior knowledge about the data, especially in noisy real-world data.

\subsection*{Tensor decomposition for data processing}

One of the most common modelings of data processing is via 2nd-order tensors (i.e., matrices). \cite{Peng:2012} transformed taxi trips into a time-zone matrix and applied a \emph{Non-Negative Matrix Factorization} (NMF)~\cite{lee1999} model to identify basic spatial and temporal patterns of human mobility. Recently, \cite{zanabria2019} designed visual analytic functionalities that allow users to select and analyze regions of interest in terms of their spatiotemporal hotspots and crime patterns based on NMF. However, in many cases, we find limitations in modeling the relationships between data with its use. Many data are generated by complex processes, which are difficult to represent without suitable structures. Hence, there is a need for new models to better understanding these data. 

The interest in applying tensors in data analysis has grown both in academia and the industry. Tensors are multi-modal arrangements capable of representing the various aspects of the data as independent modes. There are several tensor decomposition methods, such as \emph{Canonical Polyadic Decompositions} (CPD)~\cite{harshman1970}, \emph{Tensor Train Decomposition} (TTD)~\cite{Oseledets2011}, and \emph{Tucker Decompositions} (TD)~\cite{kolda2009tensor}. Each of these factorizations has different characteristics and different purposes regarding the application. For instance, the CPD has been used in signal processing and data analysis, mainly because of its facility for data interpretation~\cite{vasile2008uncovering}. TTD has been used to solve problems like stochastic partial differential equations and high-dimensional elliptic equations~\cite{Oseledets2011}. In contrast, TD has been used more frequently in applications involving data compression~\cite{cichocki2015tensor}. Recently,  Liang et al.~\cite{Liang:2022} proposed a prediction framework that explores spatial-temporal correlations in crime data using a tensor learning technique that finds the optimal solution by solving a compact optimization problem modeled with CPD.

On the order hand, in some applications is essential to keep the \emph{non-negative} property of a tensor (i.e., if all its entries are non-negative) to preserve the interpretability of data.
For that, it is necessary to introduce the non-negativity constraints, and thus we have the \emph{Non-negative Tucker Decomposition} (NTD)~\cite{cichocki2009nonnegative}.
NTD has been applied to feature extraction in many contexts. For instance, \cite{Phan:2010} proposed a model reduction and feature extraction for large-scale problems. \cite{jukic2013} proposed a feature extraction method suitable for color medical images.
\cite{Zhang2013} proposed an NTD-based framework to extract urban mobility patterns from a massive taxi trajectory data set in Beijing
Analogously, \cite{Sun:2016} applied an NTD model to decompose high-dimensional mobility data into a meaningful pattern. \cite{wang2022} provided a framework to reveal the spatial-temporal patterns of urban mobility by exploring massive and high-dimensional mobile phone data.

Despite all the works described above using tensor decomposition to extract the patterns, they are not deal straightforwardly with urban data from multiple data sources since the urban patterns require analysis and comprehension of a combination of each mode's significant feature.

\section{Data Sets}\label{data}

Understanding the urban patterns from S\~ao Paulo city has long been a research interest topic for experts. According to them, pattern extraction should rely on a mechanism to extract hidden relevant patterns from multiple data sources to cluster places with similar patterns. For~this purpose, we interact with a renowned group of sociologists from the Nucleus of Study of the Violence at the University of S\~ao Paulo (NEV-USP)~\cite{NEV}. One of the sociologists has extensive experience in sociology, emphasizing violence, urban and social conflicts. The other sociologist is an expert in geo-information and sociology applied to spatial analysis, urban planning, public security, murder, and criminal dynamic. In partnership with the police department of the state of S\~ao Paulo, the team of experts built a data set containing twelve years of urban records. 

\myparagraph{Nomenclature.} Before further detailing the problem, we define some nomenclature that will be employed throughout the manuscript.

\noindent -- \textbf{\emph{Site}}  the spatial location of an actual or planned structure or set of structures (such as a building, residences, schools, or monuments).


\noindent -- \textbf{\emph{Urban Pattern}}  refers to a set of sites with similar urban, infrastructure and criminal characteristics. 
For instance, an urban pattern would be a group of schools with high social urban class, far from bars but near bus stops and high numbers of passerby crimes at night. 

\noindent -- \textbf{\emph{Relevant Patterns}} refer to hidden correlated patterns in the data set, which are not easy to find. It is important to highlight that the founded patterns should also be representative (those with the largest number of sites possible). 

\noindent -- \textbf{\emph{Target}} refers to a site of interest in which the user wishes to figure out the patterns around.

\noindent -- \textbf{\emph{Crime type}} refers to the type of criminal activity. In particular, we have considered passerby, commercial establishment, and interior vehicle robberies.


\noindent -- \textbf{\emph{Characteristics}} refer to data sources gathered by the domain experts, such as bus stops, crime types, urban social classes, etc. 

\begin{table}[t]
    \centering
	\caption{This table summarizes the data set used during the analyses. One is composed by five columns: the type of data set, period which the data set ranges, the categories which the data set was divided, the source of the data and their availability, and the amount of data.}
	\begin{tabular}{cclccc}
		\toprule
		\rowcolor{PastelBlue}\textbf{Data Set} & \textbf{Years} & \textbf{Categories} & \textbf{Source} & \textbf{Public} & \textbf{Amount}\\ 
		\hline
		\rowcolor[gray]{.9}Crime& 2006-2017 & \begin{tabular}[c]{@{}l@{}} $\bullet$ Passerby\\ $\bullet$ Commercial estab.\\ $\bullet$ Vehicle\end{tabular} & \begin{tabular}[c]{@{}c@{}}\cite{NEV} \end{tabular} & No & \begin{tabular}[c]{@{}c@{}} 1013400 \\ 6889 \\ 81299 \end{tabular} \\
		Social Indicators  & 2010 & \begin{tabular}[c]{@{}l@{}} $\bullet$ Socio-economics\\ $\phantom{\bullet}$ ({\sf A-B,C,D,E-F-G,H})\\ $\bullet$ Homicides\\
			$\phantom{\bullet}$ (none, low, high, others)\\
		\end{tabular} & \cite{NEV} & Yes & 18953 \\
		\rowcolor[gray]{.9} Infrastructure & \begin{tabular}[c]{@{}c@{}} 2016 \\ 2011-2017  \end{tabular} & \begin{tabular}[c]{@{}l@{}}$\bullet$ Bus stops\\ $\bullet$ People flow\end{tabular} & \begin{tabular}[c]{@{}c@{}}\cite{CEM}\\ \cite{portalSP} \end{tabular} & Yes &\begin{tabular}[c]{@{}c@{}} 99856 \\ 400M  \end{tabular} \\
		%
		\bottomrule
	\end{tabular}
	\label{tab:my-table}
\end{table}

\subsection{S\~ao Paulo data}\label{data-dataset}

The data set assembles several types of data sources, as shown in Table~\ref{tab:my-table}. The police department of S\~ao Paulo provided criminal records, and only criminal acts as to robbery were provided, leaving out drug-related felonies and sexual assault. Each record contains the crime's geographic coordinates, the type of crime, the date and period (dawn, morning, afternoon, and night) of the crime occurrence. 
The data set contains crime records from 2006 to 2017 and three types of crimes: passerby, commercial establishment, and vehicle robberies. However, we decided to include only information from 2011 to 2017 in our studies because of the data set range like people flow. The~experts also created urban indicators for some variables, such as urban expansion, mobility, housing, homicides, conditions, and changes in the population and environment. In this way, each indicator assigned a label to the sites. The social indicators were also used to create a social pattern that classifies the sites from S\~ao Paulo into eight different categories, ranging from social class~{\sf A} (where the population has a high level of basic sanitation) to social class~{\sf H}  (where the population has the worst level of basic sanitation). Since there are no homicides records in our data set, we employ the homicides indicator provided by domain experts in our analyses.

In addition to the crime data set, our analysis considers more data sources, such as infrastructure, urban classes, and people flow. About the infrastructure, for people flow, we estimate it based on bus tickets downloaded from the website of S\~ao Paulo State Transparency Portal \cite{portalSP}. The bus stops data set was provided by the Brazilian research group called \textit{Centro de Estudos da Metr\'opole} (CEM)~\cite{CEM}.  

It is essential to highlight that the whole data set described above is georeferenced.

\section{Non-Negative Tensor Factorization to Pattern Identification}\label{sec:tensorAnalyzer}

\begin{figure*}[!t]
	\centering
	\includegraphics[width=\linewidth]{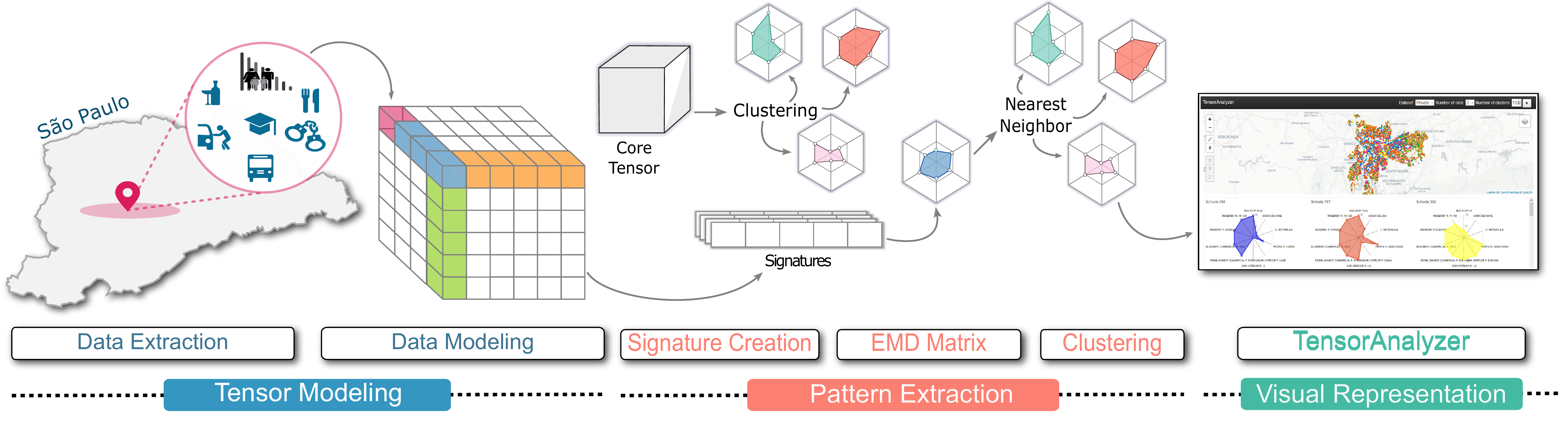}
	\caption{The overview of our entire analytical process: we started from the combination of different data sources into a tensor; after the modeling, we applied non-negative tensor factorization to extract relevant patterns; finally, the patterns are analyzed and represented using some visual resources.}
	\label{fig:pipeline}
\end{figure*}

Many data science applications generate large amounts of data with multiple features and high dimensionality for which tensors provide straightforward modeling~\cite{kolda2009tensor}. In our application, we~can summarize a specific group of records from a query using the \texttt{count} aggregate function. For~instance, a possible aggregation for a crime type could be to select all its records in a specific region and summarize them by counting. Thus, the resulting tensor from these aggregations is non-negative.

On the other hand, extracting hidden patterns from tensor data has concrete meaning only when the tensor is non-negative. In practice, non-negative factorizations are necessary when the underlying components have a physical or behavioral meaning. 
Non-negative tensor factorizations provide a powerful tool for discovering relevant patterns in large multidimensional data sets, affording compact data representations (data reduction), and preserving the data's interpretability (non-negativity).

The proposed methodology for urban pattern detection around a site using non-negative tensor factorization comprises three main steps (see Fig.~\ref{fig:pipeline}): tensor modeling, pattern extraction, and visual representation.
Before describing our methodology, we introduce some basic concepts about non-negative tensor factorization. 

\subsection{Non-Negative Tucker Decomposition}

Among the tensor factorizations,  Non-Negative Tucker Decomposition (NTD)~\cite{cichocki2009nonnegative} can be considered a higher-order generalization of NMF, a successful method for detecting fundamental features of the data~\cite{lee1999}.

Mathematically, a $n$th-order tensor $\T{X} \in \R^{I_{1} \times I_{2} \times \cdots \times I_{n}}$ is a multidimensional array, where an~element of~$\T{X}$ is indexed by a $n$-tuple of indices $(i_1,i_2,\ldots,i_n)$ and denoted by $x_{i_1 i_2 \ldots i_n}$. 
Each indexed dimension~$i_{k}$ is  called a \emph{mode} of the tensor and ranges from~1 to $I_k$. The value $I_k \in \N$ is the \emph{size} of the $k$-mode.
For~instance, a~first-order tensor is a vector, a second-order tensor is a matrix, and tensors of order three or higher are called higher-order tensors. 

Matrices and vectors can be multiplied by tensors. For our purposes, we~use the  \emph{k-mode matrix product}, i.e., a multiplication along mode $k$ of a tensor with a matrix.
Consider a tensor $\T{X} \in \R^{I_1 \times \cdots \times I_n}$ and a matrix $\V{A} \in \R^{I_k \times J}$, the $k$-mode matrix product, denoted by~\aspas{$\times_k$}, is (elementwise) defined by:
$$
(\T{X} \times_k \V{A})_{i_1 \ldots i_{k-1} j i_{k+1} \ldots i_n } = \displaystyle \sum_{i_k=1}^{I_k} x_{i_1 \ldots i_n} a_{ji_{k}} \,.
$$
Furthermore, analogous to the matrix Frobenius norm, a tensor norm  is given by:
$$
\| \T{X} \|_F = \displaystyle \sqrt{ \sum_{i_1=1}^{I_1} \sum_{i_2=1}^{I_2}  \cdots  \sum_{i_n=1}^{I_n} x_{i_1 i_2 \ldots i_n}^2 } \,.
$$ 

\begin{figure*}[!t]
	\centering
	\includegraphics[width=0.5\linewidth]{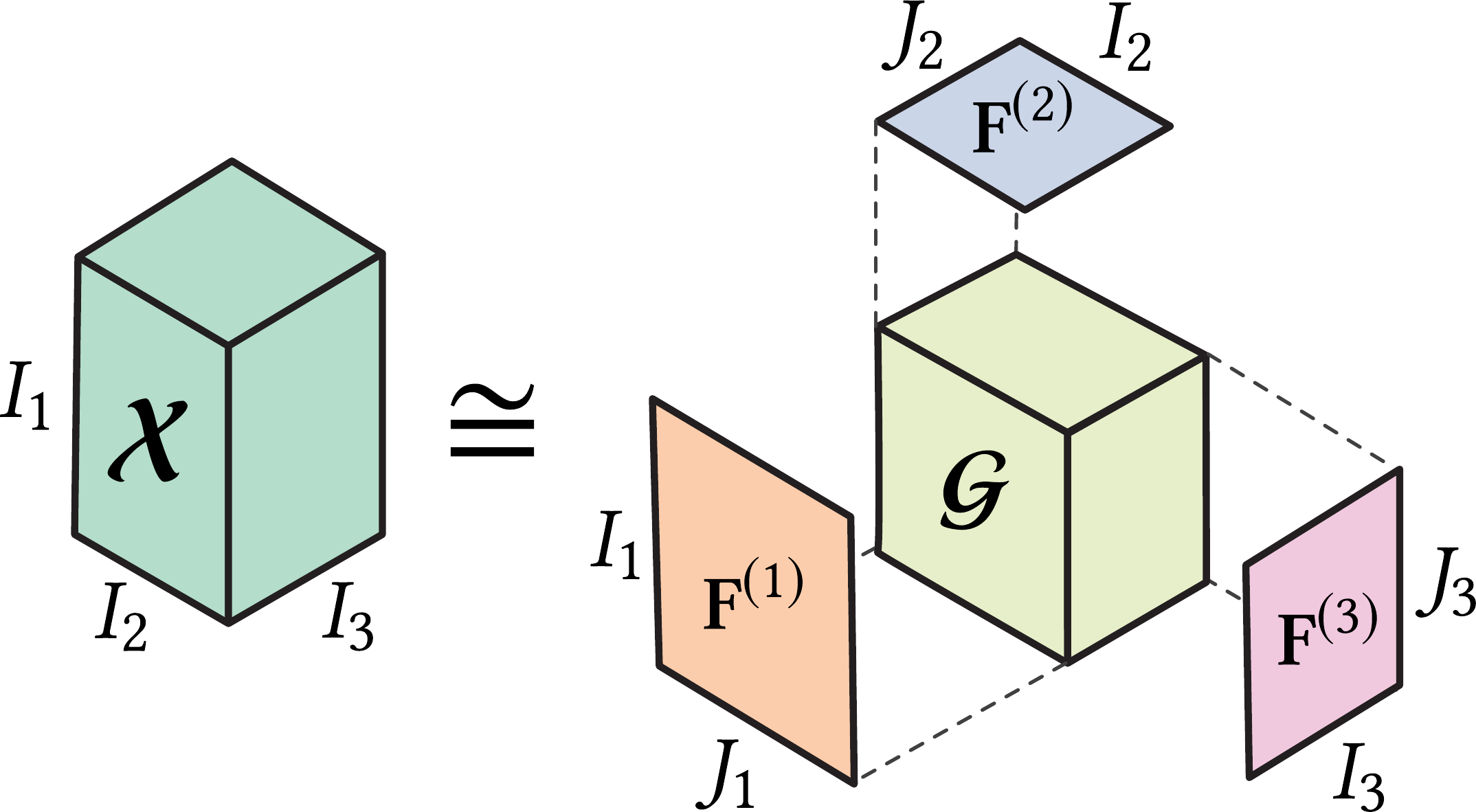}
	\caption{Illustration of a NTD for a 3rd-order tensor $\T{X} \in \R^{I_1 \times I_2 \times I_3}$. The main goal is to find the optimal factor matrices $\V{F}^{(k)} \in \R^{I_k \times J_k}$ with $k=1,2,3$ and a core tensor $\T{G} \in \R^{J_1 \times J_2 \times J_3}$, typically $J_k \ll I_k$. }
	\label{fig:tucker3}
\end{figure*}

Given a non-negative tensor $\T{X}$, NTD searches a non-negative \emph{core tensor} $\T{G} \in \R ^{J_1 \times \cdots \times J_n} $ and non-negative \emph{factor matrices} $\V{F}^{(1)} \in \R^{I_1 \times J_1},\ldots,\V{F}^{(n)} \in \R^{I_n \times J_n}$, such that 
\begin{equation}
\T{X} \cong \widehat{\T{X}} \quad \text{with} \quad  \widehat{\T{X}}  = \T{G} \times_1 \V{F}^{(1)} \times_2 \cdots \times_n \V{F}^{(n)} \,.
\label{eq:ntd}
\end{equation}
The \emph{mode ranks} $J_k$ of the core tensor are usually chosen smaller than the mode ranks $I_k$ of the original input tensor. Fig.~\ref{fig:tucker3} shows a NTD of a 3rd-order tensor.

In fact, the NTD is achieved by solving the optimization problem:
$$
\min_{\widehat{\T{X}} }  \, \frac{1}{2}  \| \T{X} - \widehat{\T{X}}  \|^2_F \,, \quad \text{subject to} \quad \T{G} \geq 0,  \V{F}^{(1)} \geq 0, \ldots,  \V{F}^{(n)} \geq 0 \,.
$$
The core tensor $\T{G}$ is a compressed version of~$\T{X}$ due to sparsity introduced by the non-negative constraints. 
Efficient algorithms to solve the optimization problem can be found in~\cite{Friedlander:2008,Xu2013}.

Each feature (factor) matrix $\V{F}^{(k)}$  comprises a piece of latent information within the original tensor data~$\T{X}$.
More precisely, we can rewrite the Equation~\eqref{eq:ntd}, as follows:
\begin{equation}
\T{X} \cong \displaystyle \sum_{j_1=1}^{J_1} \sum_{j_2=1}^{J_2}  \cdots  \sum_{j_n=1}^{J_n} g_{j_1 j_2 \ldots j_n} \, \V{f}^{(1)}_{j_1} \circ \V{f}^{(2)}_{j_2} \circ \cdots \circ \V{f}^{(n)}_{j_n} \,,
\label{eq:ntd2}
\end{equation}
where $\V{f}_{j}^{(k)}$ is the $j$-th column of the $\V{F}^{(k)}$ and the symbol \aspas{$\circ$} denotes the outer product. The entries of the core tensor $g_{j_1 j_2 \ldots j_n}$ measure the level of interaction between the different vectors $\V{f}_{j}^{(k)}$ within the associated feature matrices.
In other words,  the entries of $\T{G}$ define how the features along each feature matrix~$\V{F}^{(k)}$  are mixed to approximate the input tensor~$\T{X}$.

\subsection{Data tensor modeling}

Modeling the input data tensor from sites is the first step of our methodology. We start grouping the sites into non-overlapped circular regions of interest (ROIs) centered at each target location. Each ROI's characteristic is aggregated by counting the number of crimes and infrastructure facilities or determining the most common urban indicators. 
The ROI radius is a user parameter; in our analyses, we use a radius of 200 meters.  

\begin{figure}[!t]
	\centering
	\includegraphics[width=\linewidth]{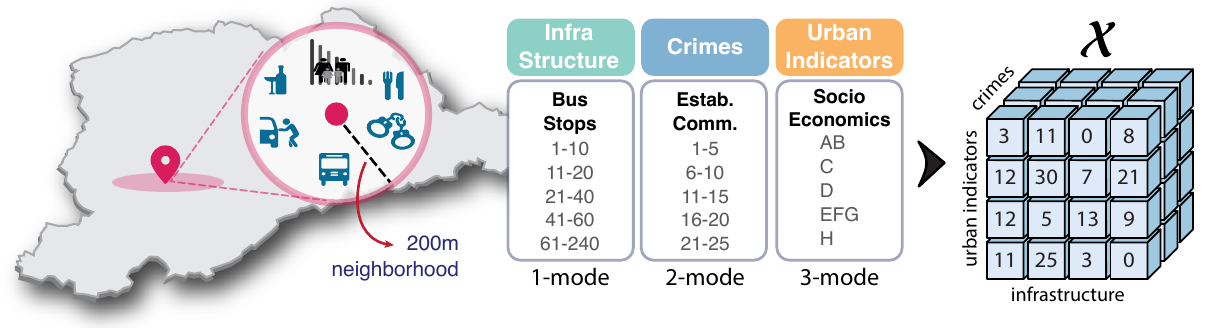}
	\caption{Data tensor modeling with 3 modes.  We create a circular ROI centered in each target location with a radius of 200~meters (\emph{left}).  We count the number of crimes, social classes, and infrastructure facilities in each ROI (\emph{middle}).  The entries of the resulting non-negative tensor $\T{X}$ are the number of sites with share the same characteristics~(\emph{right}).}
	\label{fig:tensorModeling}
\end{figure}

The aggregations result  in a 10th-order data tensor, where its modes are organized as follows: 
\begin{itemize}
	\item six modes from crime data: number of crime records (passerby, commercial
	establishment and vehicles) and its most frequent period;
	\item two modes from urban data: a most prevalent urban indicator (socio-economics and homicides);
	\item two modes from infrastructure data: number of infrastructure facilities (bus stops and people flow).
\end{itemize}
To reduce the mode sizes related to crime types and infrastructure, we use equalized histograms, i.e., the mode size is the number of histogram bins. 
The bin intervals and the categorical values are converted to numerical values (IDs) through a label encoding function $\varphi_k(i)$, which retrieves the ID associated with the $i$-th feature from the~$k$-mode. For instance, Fig.~\ref{fig:tensorModeling} shows the modeling of a 3rd-order data tensor from a site, where $\varphi_3(2)$ is the ID of the socio-economics class {\sf C}.
Finally, the~entries of the data tensor~$\T{X}$ are the number of sites with the same data characteristics, as~illustrated by~Fig.~\ref{fig:tensorModeling}.

\subsection{Patterns extraction}
After creating the data tensor~$\T{X}$,  we extract the data patterns from its NTD given by Equation~\eqref{eq:ntd}. 
The core tensor $\T{G} \in \R ^{J_1 \times \cdots \times J_n}$ reveals meaningful patterns, where the magnitude of its entries is directly related to the importance of the features encoded by the factor matrices $\V{F}^{(k)}$ and their interaction. Firstly, we extract the relevant patterns from data using the core tensor $\T{G}$. Then, we~assign each pattern of~$\T{X}$ to its nearest neighbor pattern of~$\T{G}$.

Let $g_{j_1 \ldots j_n}$ be a non-zero entry of~$\T{G}$, we~create a pattern signature as a feature vector using the most significant attribute of the columns~$\V{f}^{(k)}_{j_k}$, as follows:
\begin{equation}
\V{s}_{\T{G}} = \left(\varphi_1(i_1),\ldots,\varphi_k(i_k),\ldots,\varphi_n(i_n)\right)^{\top}  \quad \text{with} \quad i_k = \arg \max_i f_{ij_k}^{(k)}\, .
\label{eq:signatures}
\end{equation}
However, the core tensor provides many similar patterns. Thus, it is necessary to group and reduce these patterns into a small set.  
The pattern signatures are partitioned into $M$ highly related clusters using \emph{Agglomerative Hierarchical Clustering} (AHC) algorithm~\cite{Rafsanjani2012} where the distance matrix is computed using \emph{Earth Mover’s Distance}~(EMD)~\cite{Zen2014} as a similarity measure which takes into account the semantic between the signatures. 
Finally, for each cluster $C$, the signature $\V{s}_{\T{G}} \in C$ that represents the entire cluster is the one that minimizes $\sum_{\V{s} \in C} d_{\mathrm{EMD}}(\V{s},\V{s}_{\T{G}})$, where $d_{\mathrm{EMD}}$ is the~EMD.
In simple words, the pattern signature $\V{s}_{\T{G}}$ is the centroid of $C$ in the EMD sense.
%
%
%
In~our application, the number of clusters $M$ is a user parameter. Moreover, the mode ranks of~$\T{G}$ are encompassed into a single parameter~$J \in \N$ (i.e., $J_1 = \cdots = J_n = J$) also defined by the user.

Next, we need to create the pattern signatures from the input tensor~$\T{X}$ and assign them to a pattern $\V{s}_{\T{G}}$. For each non-zero entry $x_{i_{1} \ldots i_{n}}$ of~$\T{X}$, its signature is achieved straightforwardly by applying $\varphi_k$ in each mode~$k$, i.e., $\V{s}_{\T{X}} = \left(\varphi_1(i_1),\ldots,\varphi_n(i_n)\right)^{\top} $. Note that signature $\V{s}_{\T{G}}$ can be shared with many sites.

Finally, we assign $\V{s}_{\T{X}}$ to the most similar pattern of the core tensor $\T{G}$. For this purpose, we take its nearest neighbor $\V{s}_{\T{G}}$ in terms of EMD.

\begin{figure}[t]
	\centering
	\includegraphics[width=\linewidth]{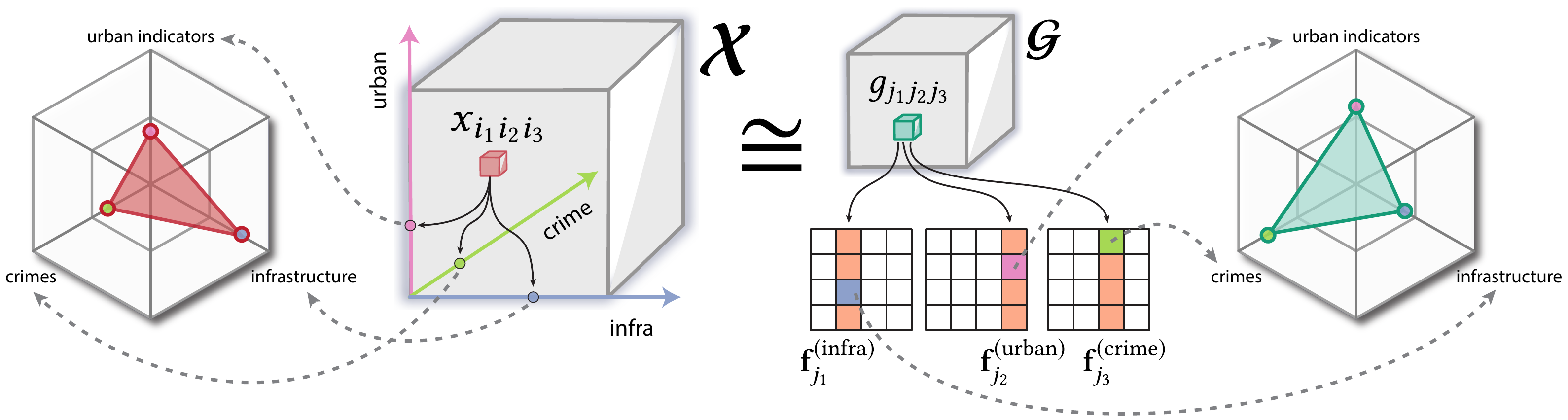}
	\caption{Visual pattern representation from tensors. The patterns of the input tensor $\T{X}$ are represented straightforwardly by a radar chart (\emph{at leftmost}) where each anchor point is defined by the signature $\V{s}_{\T{X}}$. While the entries of the core tensor $\T{G}$ provides the best combination of factors from each feature matrix. These factors are combined to form a signature~$\V{s}_{\T{G}}$ also represented by a radar chart (\emph{at rightmost}).}
	\label{fig:tensor_inte}
\end{figure}

\begin{figure*}[t!]
	\centering
	\includegraphics[width=\linewidth]{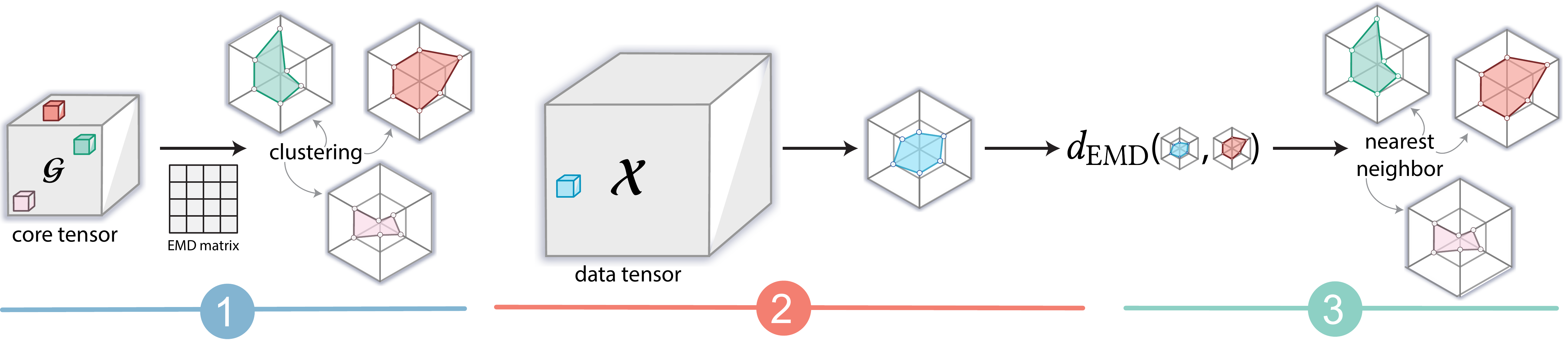}
	\caption{Patterns extraction process: (1)~the most relavant patterns returned by $\T{G}$ are clustered , (2)~patterns are extracted from $\T{X}$,  and  (3)~each pattern of $\T{X}$ is assigned to its nearest pattern of $\T{G}$ in terms of EMD.}
	\label{fig:PatterExtraction}
\end{figure*}

The~patterns are represented visually by a simple \emph{radar chart} where each anchor point represents a mode of the tensor, as illustrated by Fig.~\ref{fig:tensor_inte}. The values in each mode define the polygon's shape, producing a visual identity for the pattern signature. 
The entire pattern extraction process is summarized by Fig.~\ref{fig:PatterExtraction}. 

\subsection{Visual design}\label{sec:visualdesign}

In this section, we describe the visual components of TensorAnalyzer. Fig.\ref{fig:teaser} illustrates the web-based system, which comprises a \textit{Control Menu} and two viewports. The domain experts assist us in the design of each visual resource. \textit{Map View} shows the geolocation of the target locations and their associated pattern, while \textit{Patterns View} shows the patterns (radar chart) extracted by our methodology. 

Visualization techniques have been successfully employed in crime analysis to explore the data~\cite{zanabria2019,Liang:2022}. However, new designs over existing visualizations are required. For instance, the \textit{Patterns View} is a novel alternative visualization in this context, which turns out to help elucidate the relationship between crime and other characteristics (urban indicators and infrastructure facilities) involved in the analysis. Although the information visualization community knows this visual metaphor well, to our best knowledge, it has never been used for urban data analysis. In the following paragraphs, we describe each visual component.

\myparagraph{Control Menu.} This toolbar provides the following user selections: the input data set, the mode rank~$J$ for the core tensor for the NTD step, and the number of pattern clusters~$M$. 

\myparagraph{Map View.} Each point represents a target location with its geolocalization, and its color corresponds to the associated pattern. This visualization allows the users to identify and explore which sites are in a specific pattern.

\myparagraph{Patterns View.} This viewport allows the user to visualize all patterns detailed and the number of schools with the same characteristics. Moreover, it is possible to select a specific pattern and visualize just the sites which belong to it.

\section{Results}

\begin{figure}[t!]
	\centering
	\subfigure[Random patterns from a normal distribution around the target patterns A1, A2 and A3: 500 samples for each pattern with  10\%, 13\% and 15\% of noise for A1, A2 and A3, respectively.]{
		\includegraphics[width=.85\linewidth]{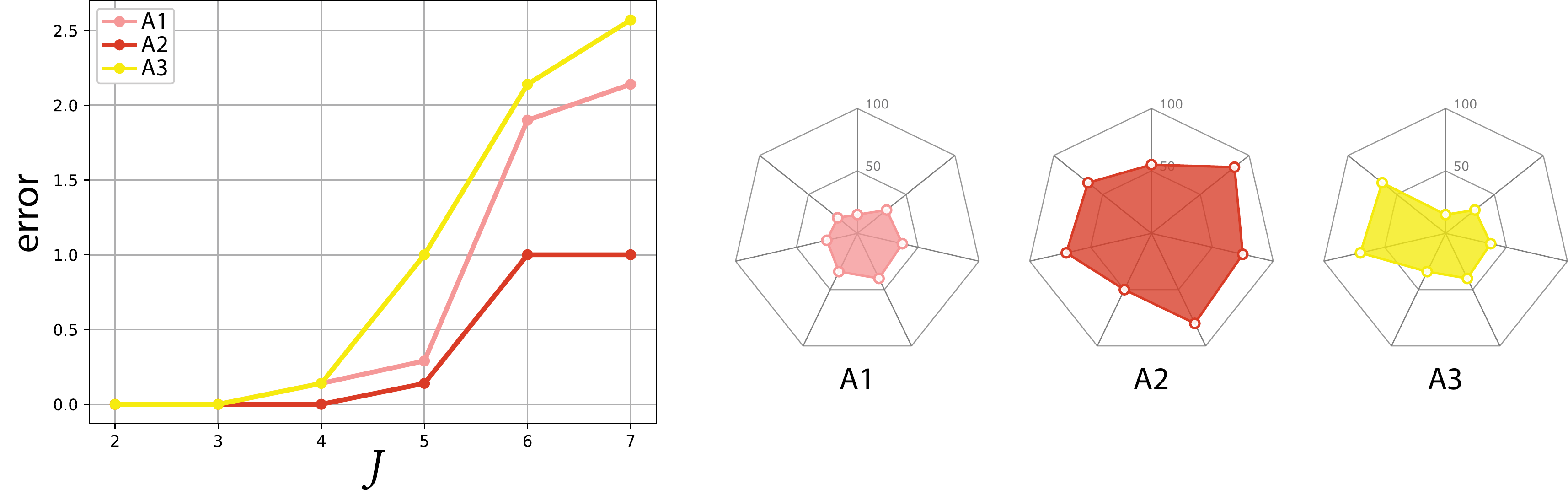}
		\label{fig:noise-P}
	}
	\subfigure[Random patterns from a normal distribution around the target patterns B1, B2 and B3: 500, 1000, 1500 samples  with 10\%~of noise for B1, B2 and B3, respectively.]{
		\includegraphics[width=.85\linewidth]{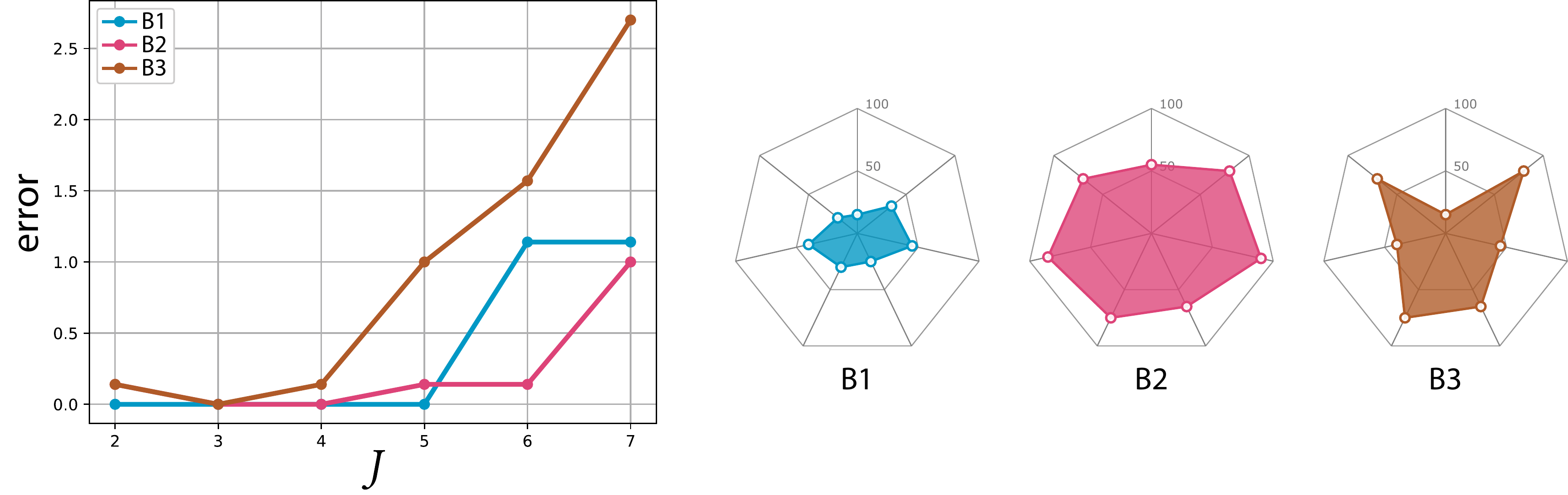}
		\label{fig:noise-S}
	}
	\subfigure[Random patterns from a normal distribution around the target patterns C1, C2 and C3: 500~samples  with 16\%~of noise for~C1,   1000~samples with 11\%~of noise for C2 and  1500~samples  with 13\%~of noise for C3.]{
		\includegraphics[width=.85\linewidth]{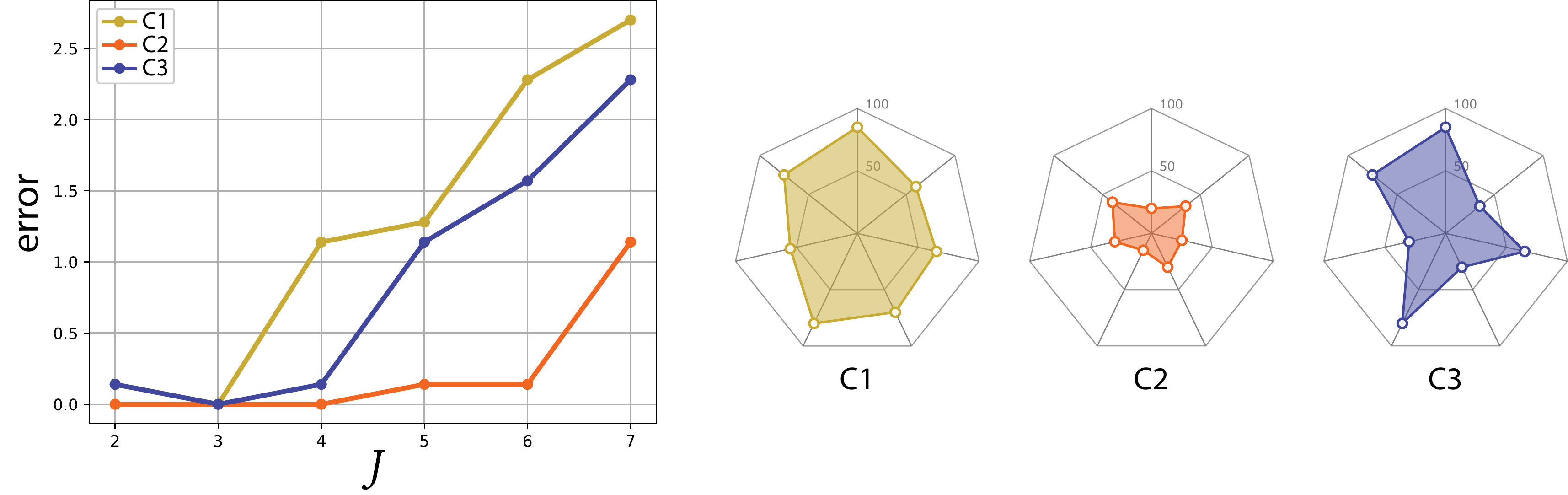}
		\label{fig:noise-M}
	}
	\caption{The error analysis of TensorAnalyzer regarding the rank $J$, the number of samples and the presence of noise in the data.}
	\label{fig:noise}
\end{figure}

To validate the effectiveness of our methodology, we present a set of comparisons using synthetic data sets and case studies from real data sets of S\~ao Paulo city. Firstly, we show the patterns identified by TensorAnalyzer from noisy data (Section~\ref{sec:noise}) and a comparison against prior feature vector clustering approaches (Section~\ref{sec:cluster}). Then, we perform case studies to verify the accomplishments of the experts presented in Section~\ref{sec:case}.

We implemented our visual interface in JavaScript using \texttt{D3}~\cite{bostock:tvcg:2011}. 
To compute the NTD, we use the \texttt{Tensorly} Python library~\cite{tensorly}. For the AHC, we use its implementation provided by the \texttt{Scikit-learn}~\cite{scikit-learn}. All~data pre-processing and experiments have been conducted on a 4-core 2.8~GHz Intel i7-4980HQ with 16GB of RAM.

\subsection{Identifying patterns with NTD}\label{sec:noise}
We rely on NTD and their robustness against noise to identify patterns. To simplify the discussion, we present the proposed approach using synthetic examples. In particular, we~draw random samples from a $n$-dimensional normal distribution $\mathcal{N}(\bm{\mu},\bm{\Sigma})$,  where the mean vector $\bm{\mu}$ is taken as a target pattern signature (ground truth) and the covariance matrix~$\bm{\Sigma}$ is the identity matrix. 

In~our experiments, we consider three clusters ($M=3$) in $\R^7$  with a different number of samples and noise intensities. This construction simulates three regions where urban data was collected, and we desire to extract the target pattern from each cluster using our NTD approach. Fig.~\ref{fig:noise} shows the error analysis regarding the rank $J$ in three different scenarios, where the error is defined by $d_{\mathrm{EMD}}$ between the target signature and the pattern provided by TensorAnalyzer. As can be seen, the low-rank values can recover target patterns from noisy data. Moreover, an initial guess for the rank is $J= \lfloor n/2\rfloor $.

\subsection{Comparison against clustering algorithms}\label{sec:cluster}

In this section, we compare our methodology against well-known clustering techniques, such as k-means and AHC (without NTD). 
In this experiment, we generated an input synthetic data set in $\R^6$ organized in ten clusters of different sizes. Thus, the observations are randomly sampled in regions $\Omega_k \subset \R^6$ and assigned to each cluster $C_k$, as detailed by Table~\ref{tab:table-toy}.

\begin{table}[h]
\centering
\caption{Synthetic data set with 10 clusters.}
\vspace*{-0.2cm}
\begin{tabular}{ccl}
\toprule
\rowcolor{PastelBlue} \textbf{cluster}  & \#\textbf{elements} & \textbf{region} $\Omega_k$ \\  \hline
\rowcolor[gray]{.9} $C_1$  & 500  & {\small $[1,10] \times [1,5] \times [1,30000] \times [1,50] \times [1,60] \times [1,7]$} \\
$C_2$  & 600  & {\small $[11,20] \times [6,10] \times [31000,60000] \times [51,70] \times [61,90] \times [8,11]$} \\
\rowcolor[gray]{.9} $C_3$  & 700 & {\small $[21,30] \times [11,15] \times [61000,80000] \times [71,90] \times [91,120] \times [12,16]$} \\
$C_4$  &  800  & {\small $[31,40] \times [16,21] \times [81000,100000] \times [91,120] \times [121,150] \times [17,21]$} \\
\rowcolor[gray]{.9} $C_5$  & 900 & {\small $[41,50] \times [22,30] \times [101000,130000] \times [121,150] \times [151,180] \times [22,27]$}\\
$C_6$  & 10  &  {\small $[51,60] \times [1,5] \times [61000,80000] \times [91,120] \times [121,150] \times [22,27]$}\\
\rowcolor[gray]{.9} $C_7$  & 10 & {\small $[21,30] \times [31,40] \times [1,30000] \times [71,90] \times [151,180] \times [17,21]$}\\
$C_8$  &  10  & {\small $[31,40] \times [11,15] \times [131000,150000] \times [121,150] \times [91,120] \times [12,16]$} \\
\rowcolor[gray]{.9} $C_9$  & 10 & {\small $[41,50] \times [1,5] \times [101000,130000] \times [151,180] \times [61,90] \times [17,21]$} \\
$C_{10}$  &  15  & {\small $[1,10] \times [6,10] \times [61000,80000] \times [91,120] \times [121,150] \times [22,27]$} \\
\bottomrule
\end{tabular}
\label{tab:table-toy}
\end{table}

Considering the knowledge of the ground truth, i.e., knowing the cluster labels in advance, we compared both approaches by using traditional measures to evaluate the clustering performance~\cite{scikit-learn}: \emph{Fowlkes-Mallows index}~(FMI),  \emph{Adjusted Rand Index}~(ARI), \emph{V-measure}, and \emph{Mutual Information} (MI).
Although the measures are derived from the confusion matrix, they have different meanings. For instance, FMI and ARI are built on counting pairs of objects classified similarly in both clusterings. On the other hand, V-measure and MI are based on conditional entropy analysis.
The scores of all measures range from 0.0 to 1.0, and higher scores lead to better results.

\subsection{Case study: criminality around São Paulo's schools} \label{sec:case}

Analyzing patterns extracted from urban data from school neighborhoods allows researchers and public policymakers to understand the relationship and the interplay between crimes, socioeconomic indicators, urban infrastructure, and school types. In our case, school type refers to whether the school administration is public (state and municipal) or private.

The type and pattern of crimes change considerably around S\~ao Paulo city, which presents regions where crimes occur with high frequency and are typically accompanied by gratuitous violence. Understanding the connection between crime patterns and the characteristics of each region has long been a topic of research interest~\cite{de2018spatial}. Several studies have been accomplished to explain how socioeconomic variables (population, rent values, economic level, and unemployment rate) and urban infrastructure (presence of bars, banks, and schools) affect particular crime types~\cite{caplan2010risk,gomez2012statistics}. 

Specifically, the study of criminal activities near schools has also been of great interest since 46.8\% of the state schools of S\~ao Paulo reported at least one case of violence between January 2007 and May 2009. In this period, it was recorded $27,340$ occurrences which range from acts against the patrimony to acts against people~\cite{TAVARES2016}. Roncek and LoBosco~\cite{roncek1983effect} accomplished a study showing that the neighborhood of high schools tends to present higher rates of crimes than other regions with similar social and urban characteristics. In contrast, the study performed by Murray and Swatt~\cite{murray2013crime} indicates that the presence of schools tends to reduce the number of robberies in their neighborhoods. Those conflicting studies suggest that patterns and crime types around schools vary substantially depending on the city and its properties. Therefore, analyzing criminal activities around schools in different cities is imperative to determine whether those cities share specific crime patterns and what factors drive such a similarity.

In this context, we run several rounds of meetings and interviews with the experts to specify the main problem requests involved in analyzing criminal data. The outcome of such interactions can be summarized as follows:

\begin{description}
	\item[R1] \textbf{-- Understanding the relations between crime events and the other variables involved in the analysis near schools.} 
	The effects of urban characteristics on crime in Latin American cities are little studied \cite{montolio2018}. The understanding of this relationship could inform urban planning that helps deter crime. For instance, quantifying crime variations and their relationship with population flows through the different bus stations could help policymakers understand the impact of urban infrastructure modifications on crime near schools. \cite{block2000} showed the effect of some variables on criminality increase in the Bronx, and one of their findings is that the subway stop can increase four times the robberies. Based on their study, experts would like to elucidate whether there is a relationship between crime and bus stops near schools. 
	
	\item[R2] \textbf{-- Analysis of how criminality influences the students' performance.}
	The study conducted by researchers at Northwestern, New York University, and DePaul University \cite{heissel2018}, found that violent crime changes the sleep patterns of students living nearby, which increases the amount of the stress hormone cortisol in the students' bodies the day immediately following the violent incident. Sleep disruption and increased cortisol have been shown to impact students' performance in school negatively. In this context, experts would like to investigate whether violent crimes have affected the students' performance in S\~ao Paulo and what patterns represent this relationship.
	
	\item[R3] \textbf{-- Understanding criminal patterns of recreational areas near schools.}
	From the expert's experience, parks and recreational areas are regions of severe crimes, including assaults, thefts, and sexual assaults. Although assaults and thefts are expected during the day, experts believe higher crime rates are reported during nighttime. So, investigating the patterns near areas frequented by visitors is challenging. 
\end{description}

\begin{figure}[t!]
	\centering
	\subfigure[Private schools.]{
		\includegraphics[width=.85\linewidth]{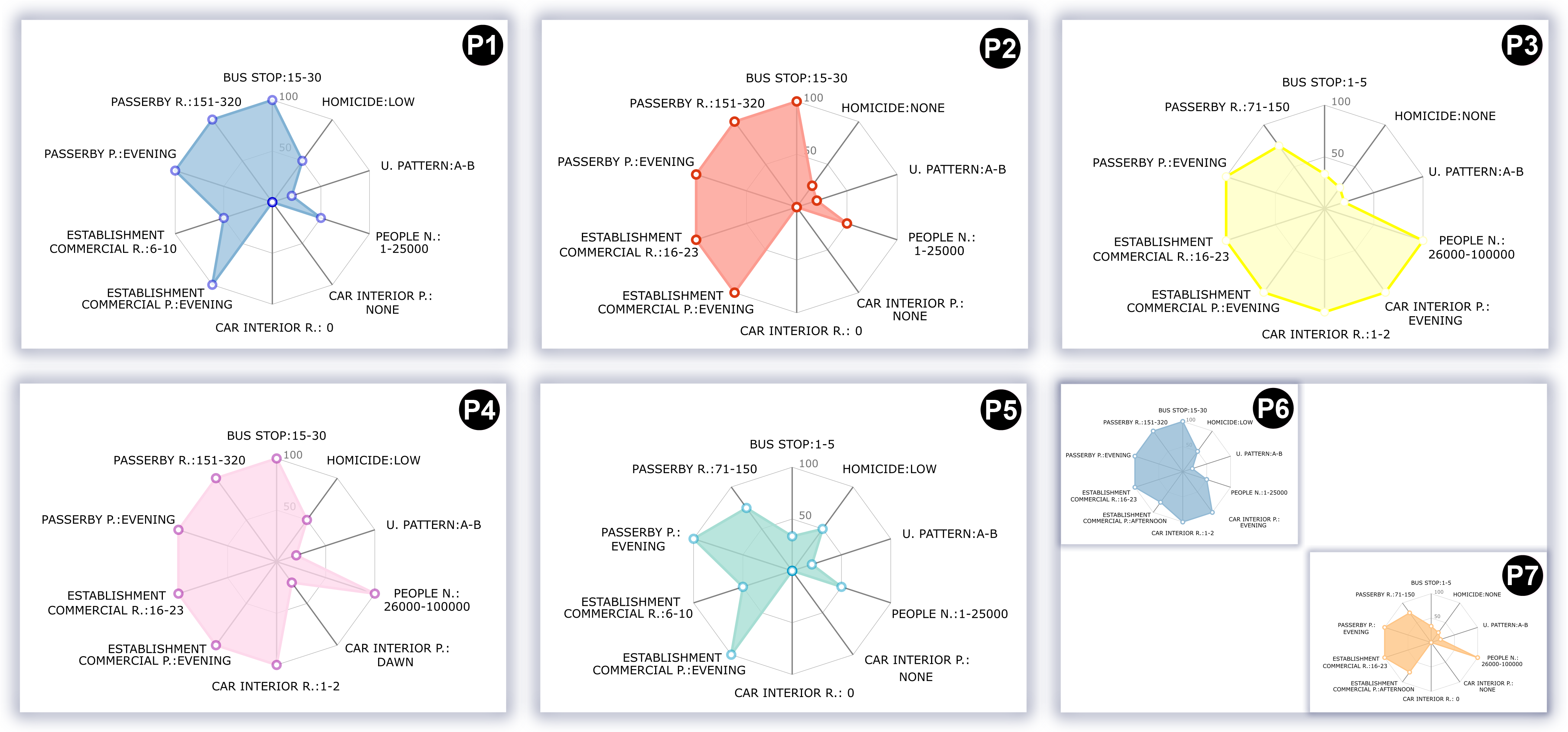}
		\label{fig:patterns-infrastructure-private}
	}
	\subfigure[State schools.]{
		\includegraphics[width=.85\linewidth]{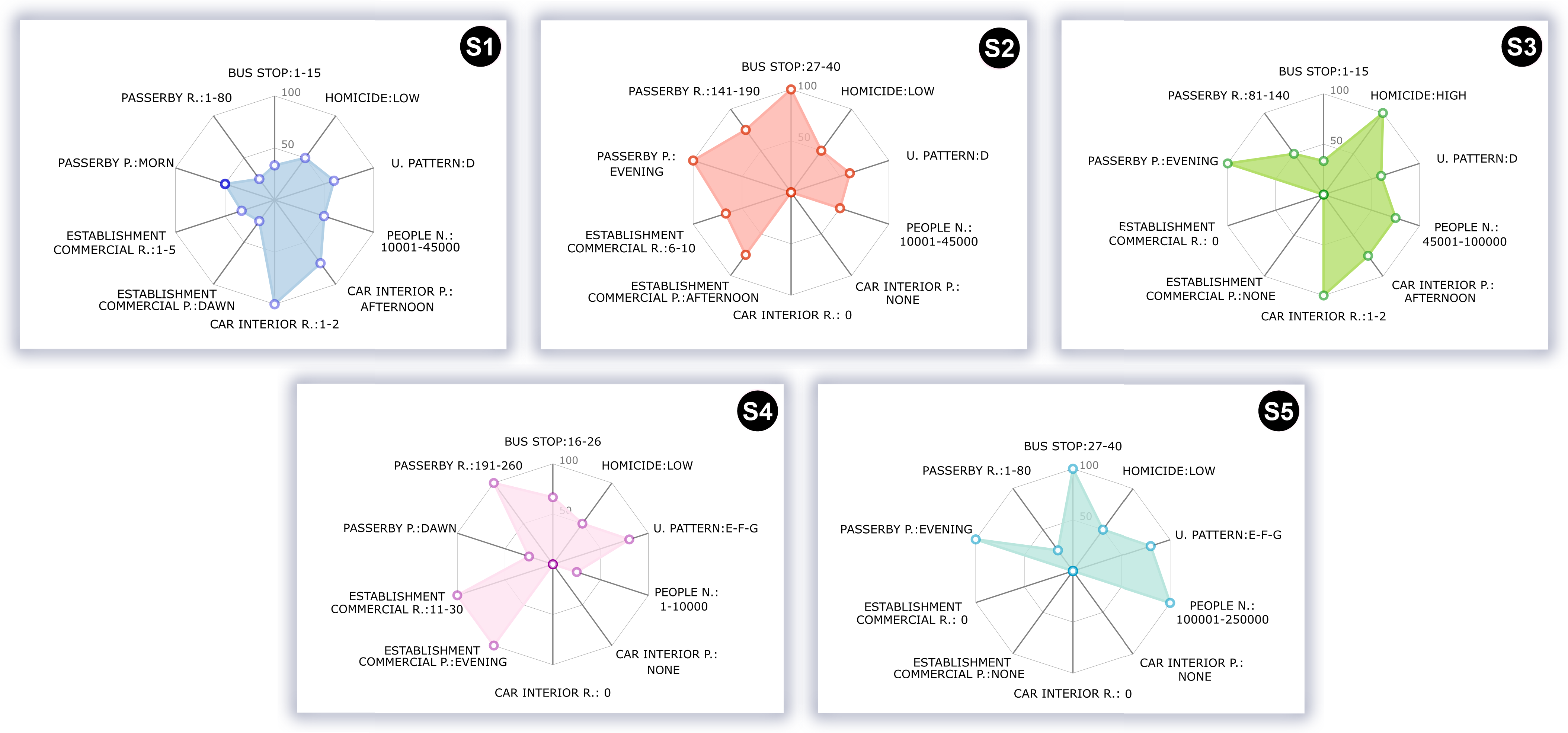}
		\label{fig:patterns-infrastructure-state}
	}
	\subfigure[Municipal schools.]{
		\includegraphics[width=.85\linewidth]{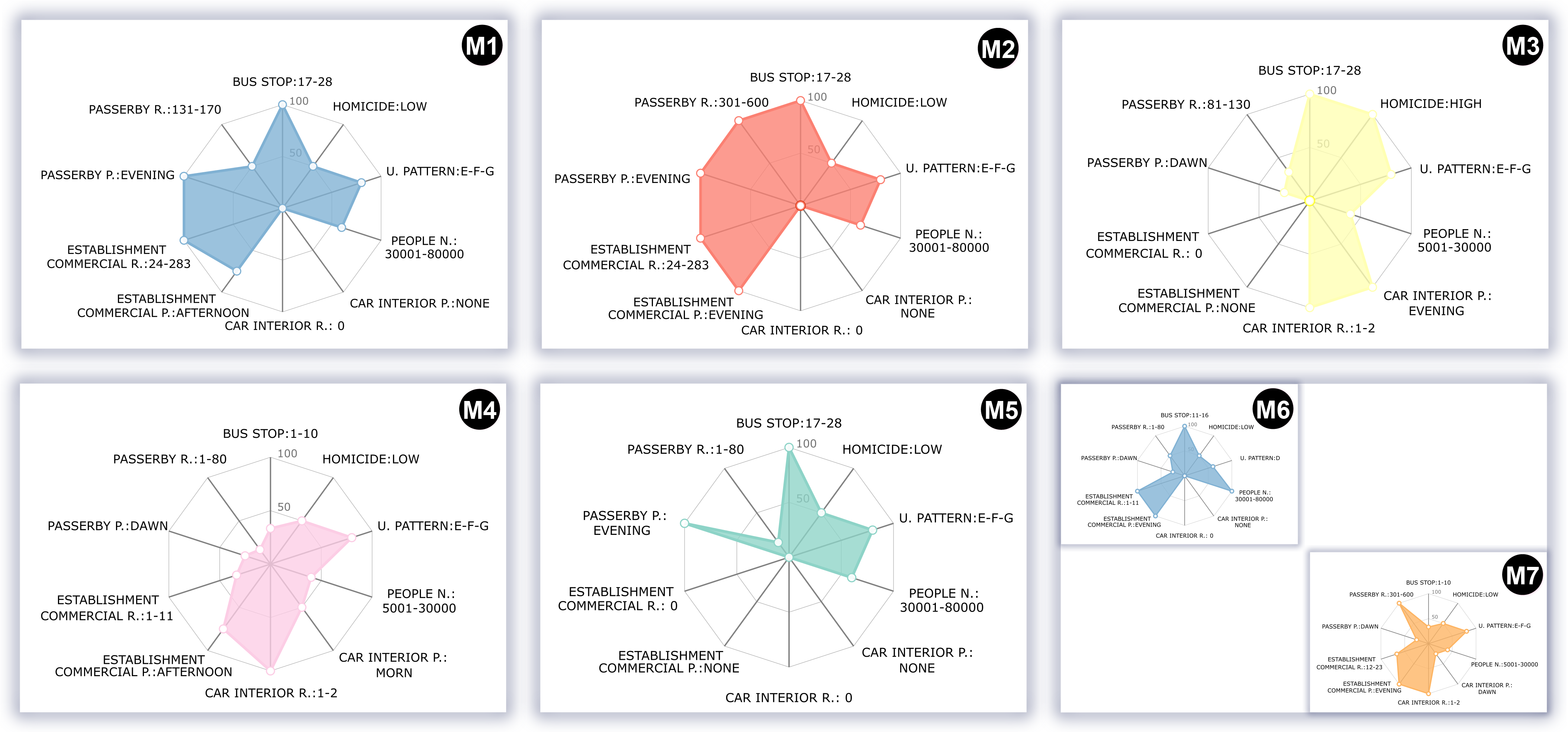}
		\label{fig:patterns-infrastructure-city}
	}
	\vspace*{-0.3cm}
	\caption{Patterns provided by TensorAnalyzer with rank $J = 5$.}
	\label{fig:patterns-school}
\end{figure}

Therefore, this case study aims to understand the relations between crime patterns, students' performances, and the characteristics in the schools' surroundings of S\~ao Paulo. More precisely, we want to understand how crimes affect the students' performances, how infrastructure can influence the crime increment, and the prevalent crime pattern of the frequented visitors. To~perform these analyses, we use the school data set provided by CEM~\cite{CEM}.
The schools' data set has 11931 records, and each one contains information about the type of the school (private, state, or municipal), as can be seen in Table~\ref{tab:schools}, and students' performance indicators ranging from 2012 to 2015, such as:

\begin{itemize}
	\item \textit{National High School Exam} (NHSE) is an exam applied to all students from Brazil to evaluate the High School quality and choose the students that are going to enter the public universities;
	\item \textit{Elementary Schools Educational Development Index in the Initial Years} (ESEDIY) and \textit{Final Years} (ISPIFY) is a national indicator to measure the quality of primary education (students from first to the ninth year). 
\end{itemize}

\begin{table}[t]
    \centering
	\caption{The school data set was used during our analyses.}
	\begin{tabular}{cclccc}
		\toprule
		\rowcolor{PastelBlue}\textbf{Data Set} & \textbf{Years} & \textbf{Categories} & \textbf{Source} & \textbf{Public} & \textbf{Amount}\\ 
		\hline
		Schools & 2016 & \begin{tabular}[c]{@{}l@{}}$\bullet$ Private \\ $\bullet$ State \\ $\bullet$ Municipal\end{tabular} & \cite{CEM} & Yes & \begin{tabular}[c]{@{}l@{}} 4088 \\ 1217  \\ 1539 \end{tabular} \\
		\bottomrule
	\end{tabular}
	\vspace{-0.3cm}
	\label{tab:schools}
\end{table}

\subsubsection{The increase of infrastructure increases the crime (\textbf{R1})} 

S\~ao Paulo is the most populous city in Latin America, with crime levels higher than the worldwide average. It is characterized by unplanned and high socioeconomic inequality, yet the effects of urban characteristics on crime have not been studied. Thus, understanding this relationship could inform urban planning that helps prevent crimes. Fig.~\ref{fig:patterns-school} illustrates patterns from the schools provided by TensorAnalyzer. 

\begin{figure}[b!]
	\centering
	\includegraphics[width=\linewidth]{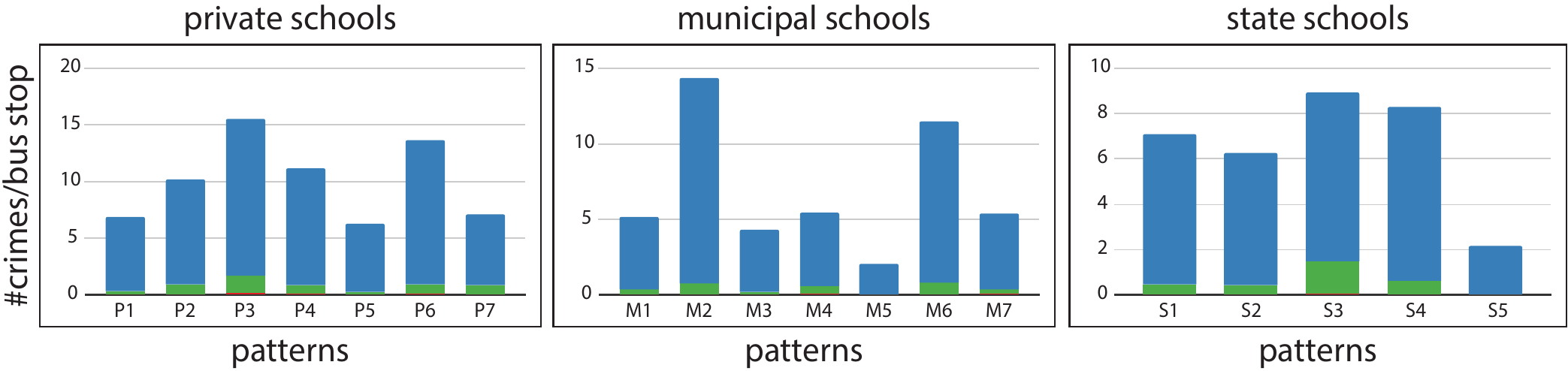}
	\caption{OLS regression from bus stops and crimes: passerby~(\crule[blue]{0.23cm}{0.23cm}), commercial establishment~(\crule[green]{0.23cm}{0.23cm}), and vehicle~(\crule[red]{0.23cm}{0.23cm}) robberies.}
	\label{fig:chart_infra_crime}
	\vspace*{-0.35cm}
\end{figure}

In order to investigate whether there is a relationship between crimes and infrastructure, i.e., we need to validate whether the increase in infrastructure facilities can influence the increase in crimes. Thus, for each data set, we apply a linear regression on the patterns to explore the relationship between the bus stops and the three types of crimes: passerby, commercial establishment, and vehicle interior. In particular, the linear regression is performed using \textit{Ordinary Least-Squares} (OLS)~\cite{Montgomery}. 
Fig.~\ref{fig:chart_infra_crime} shows a relationship between the bus stops and the crimes for each data set in which the horizontal axes represent the patterns, and the vertical axes represent the number of crimes per bus stop estimated by OLS. Passerby robbery has a strong relationship with bus stops in all data sets. For instance, the pattern M2 shows that the increase of one bus stop increases the passerby robbery 14 times in municipal schools' neighborhoods. On the other hand, we can notice that commercial establishment and vehicle robberies do not have a strong relationship with bus stops since the regression values are below one in most of the patterns. 

\subsubsection{Crime affects the students' performances (\textbf{R2})}
Crime experts would like to investigate whether violent crimes (e.g., homicides) have affected the students' performances in S\~ao Paulo and what patterns represent this relationship. 
Table~\ref{tab:performance-high} shows students' performance indicators given by NHS in the high school for private and (public) state schools. The best performance occurs in private schools where the patterns depicted in Fig.~\ref{fig:patterns-infrastructure-private} present very low or nonexistent homicide rates. Moreover, the crimes in these patterns happen in the evening, at times with no class. Otherwise, the pattern S3 from state schools in Fig.~\ref{fig:patterns-infrastructure-state} shows the highest homicide rate among the others and the worst students' performances on NHSE.

\begin{table}[!h]
    \centering
	\caption{Student's performance in high schools from the patterns shown in Figs.~\ref{fig:patterns-infrastructure-private} and~\ref{fig:patterns-infrastructure-state}. Best and worst results in bold and italic, respectively. }
	\vspace{-.2cm}
	\begin{tabular}{c|cccccccccccc}
		\toprule
		\rowcolor{PastelBlue} \textbf{Pattern} & P1 & P2 & P3 & P4 & P5 & P6 & P7 & S1 & S2 & S3 & S4 & S5 \\ 
		\hline
		\rowcolor[gray]{.9}  \textbf{NHSE} & 565 & 575 & \textbf{601} & 569 & 581 & 570 & 584 & 543 & 542 & \emph{515} & 547 & 536 \\
		\bottomrule
	\end{tabular}
	\label{tab:performance-high}
\end{table}

Table~\ref{tab:performance-elementary} shows the performance indicators given by ESEDIY and ISPIFY for students of elementary school from (public) municipal schools. Similar to the previous experiment, the pattern M3 in Fig.~\ref{fig:patterns-infrastructure-city} presents the highest homicide rate among the others and the worst students' performance in both indicators.

\begin{table}[!h]
    \centering
	\caption{Student's performance in elementary schools from the patterns shown in Fig.~\ref{fig:patterns-infrastructure-city}. Best and worst results in bold and italic, respectively. }
	\vspace{-.2cm}
	\begin{tabular}{c|ccccccc}
		\toprule
		\rowcolor{PastelBlue} \textbf{Pattern} & M1 & M2 & M3 & M4 & M5 & M6 & M7 \\ 
		\hline
		\rowcolor[gray]{.9}  \textbf{ESEDIY} & \textbf{5.65} & 5.60 & \emph{5.36} & 5.48 & 5.48 & 5.52 & 5.48\\
		\textbf{ISPIFY} & \textbf{4.45} & 4.40 & \emph{4.17} & 4.32 & 4.30 & 4.32 & 4.25\\
		\bottomrule
	\end{tabular}
	\label{tab:performance-elementary}
\end{table}

In conclusion, the experiments show a strong relationship between the homicide rate and student performance, where the performance is inversely proportional to the homicide rate. The worst results occur in poor neighborhoods of public schools with a high number of homicides.

\subsubsection{Recreational areas have experienced severe crimes, mostly at night (\textbf{R3})}

The presence of crimes in S\~ao Paulo streets is an ever-present problem, especially in the evenings and late at night. Visitors are not immune to acts of crime, and they may be susceptible to targeting certain crimes, as they tend to be more likely to
display wealth, making them a more attractive target. Parks and recreational areas frequented by visitors and citizens have experienced severe crimes. Although assaults and thefts are also common during the day, experts believe that higher rates of crimes have been reported during nighttime hours. Therefore, this analysis aims to understand the crime patterns in recreational areas and the period in which the crimes occur with more frequency. 

In private schools, most of the patterns lie in the {\sf A-B}~urban class, where we notice a lot of squares, sights, and recreational areas (frequented by visitors and citizens). Fig.~\ref{fig:patterns-infrastructure-private} shows that the passerby robbery is intense in all patterns, occurring more frequently in the evening. Therefore, visitors must be more careful in recreational areas, mainly in the evening. Moreover, public policymakers must direct an effort to reduce passerby robberies in these regions.      

\section{Discussion and Limitations}
TensorAnalyzer was developed in close cooperation with domain experts, satisfying their requirements. In this section, we highlight limitations in TensorAnalyzer and improvements that can be made in the actual state of the development and future work.

\myparagraph{Choice of parameters.} Currently, in our framework, the choice of mode rank and the number of clusters are user-defined parameters. It is necessary, some previous user knowledge of the study region. In all case studies, we use the expert's knowledge to define the parameters correctly. 
A~convenient method to select the rank of a feature automatically relies on the Fisher score~\cite{Phan:2010}. We leave the inclusion of this selection to our framework as future work.

\myparagraph{Computational time.}  Since the NTD step is computed once; we consider this step as pre-processing for TensorAnalyzer. All patterns extracted for the case study (Section~\ref{sec:case}) took 12~seconds, turning the computational time feasible for a security agent or a crime expert to perform interactive data analysis. 

\myparagraph{Data source.} We intend to add more data sources to the tensor, such as more infrastructure data. However, the infrastructure data source is not easy to manage. For instance, bars and restaurants open daily, so updating the data is challenging. Moreover, our methodology could be applied to different target locations, such as malls, hospitals, and others.

\myparagraph{New resources.} We intend to add more view resources to allow the user to perform more exploratory analysis. For instance, in some cases, crime data has information about the occurrence date. The system could allow a temporal analysis of the founded patterns, allowing experts to understand the dynamical behavior of urban patterns

\section{Conclusion}
We introduced a novel approach to extracting urban patterns distributed across multiple data sources based on the Non-negative Tucker Decomposition. In close collaboration with domain experts, we developed a generic new tool named TensorAnalyzer to support the analysis of urban patterns around a target location (which could be schools, squares, homicides, and others). 
Moreover, to validate the effectiveness of our methodology, we presented a set of comparisons using synthetic data sets and case studies from accurate data set of S\~ao Paulo city. The comparison of our methodology against well-known clustering techniques, such as k-means and AHC, showed that TensorAnalyzer outperforms the k-means and AHC, reaching higher scores in all used measures.

\section*{Acknowledgments}
This was was supported by CAPES.

\bibliographystyle{unsrt}  
\bibliography{references}

\end{document}